\documentclass[runningheads]{llncs}
\usepackage[T1]{fontenc}
\usepackage{graphicx}
\usepackage{hyperref}
\usepackage{xcolor}
\usepackage{soul}
\usepackage{booktabs}   
\usepackage{multirow}   
\usepackage{caption}
\usepackage{subcaption}
\usepackage{mwe}
\usepackage{graphicx}
\usepackage{float}
\usepackage{caption}
\usepackage{booktabs}
\usepackage{tabularx}
\usepackage{bm}  
\usepackage{tikz}
\usetikzlibrary{arrows.meta, positioning, shapes.multipart, calc}
\usepackage[misc]{ifsym}
\usepackage{natbib} 

\usepackage{tikz}
\usepackage{fontawesome5} 
\usetikzlibrary{shapes.geometric, arrows.meta, positioning}
\usetikzlibrary{shapes.geometric, positioning, arrows.meta}

\begin{document}

\title{TT-XAI: Trustworthy Clinical Text Explanations via Keyword Distillation and LLM Reasoning}

\titlerunning{TT-XAI}

\author{Kristian Miok\inst{1,2} \and
Blaz \v Skrlj \inst{3} 
\and
Daniela Zaharie \inst{4} 
\and
Marko Robnik \v Sikonja\inst{1}
}
%
%
\institute{Faculty of Computer and Information Science, University of Ljubljana, Slovenia\\
\email{\{kristian.miok,marko.robnik\}@fri.uni-lj.si}
\and
ICAM - Advanced Environmental Research Institute, West University of Timisoara, Romania 
\and
Department of Knowledge Technologies, Jo\v zef Stefan Institute, Slovenia \\
\email{blaz.skrlj@ijs.si}
\and
Department of Computer Science, West University of Timisoara, Romania \\
\email{daniela.zaharie@e-uvt.ro}}


\authorrunning{K. Miok et al.}


\maketitle              

\begin{abstract}
Clinical language models often struggle to provide trustworthy predictions and explanations when applied to lengthy, unstructured electronic health records (EHRs). This work introduces \textbf{TT-XAI}, a lightweight and effective framework that improves both classification performance and interpretability through domain-aware keyword distillation and reasoning with large language models (LLMs). First, we demonstrate that distilling raw discharge notes into concise keyword representations significantly enhances BERT classifier performance and improves local explanation fidelity via a focused variant of LIME. Second, we generate chain-of-thought clinical explanations using keyword-guided prompts to steer LLMs, producing more concise and clinically relevant reasoning. We evaluate explanation quality using deletion-based fidelity metrics, self-assessment via LLaMA-3 scoring, and a blinded human study with domain experts. All evaluation modalities consistently favor the keyword-augmented method, confirming that distillation enhances both machine and human interpretability. TT-XAI offers a scalable pathway toward trustworthy, auditable AI in clinical decision support.
\keywords{Trustworthy AI \and Clinical NLP \and Electronic Health Records \and Explainability \and Keyword Extraction \and Large Language Models}
\end{abstract}


\section{Introduction}

Deep pre-trained language models (PLMs), such as BERT family of transformer models, and their adaptations to medical texts, have become essential tools in clinical natural language processing (NLP), demonstrating strong performance across tasks like diagnosis classification, medical coding, and clinical event prediction. These models can process unstructured free-text data at scale, offering opportunities to support clinical workflows and improve patient care. However, despite their predictive power, PLMs typically lack interpretability—an essential requirement in high-stakes domains such as healthcare, where decisions must be transparent, justifiable, and auditable by medical professionals. For many medical applications, their transparency is also a legal requirement, e.g., by the EU AI Act.

To address this, post-hoc perturbation-based explanation techniques such as LIME~\citep{ribeiro2016should} and SHAP~\citep{lundberg2017unified} are used to highlight input features that influence model predictions. While effective in some domains, these methods encounter substantial limitations when applied to the lengthy, heterogeneous nature of electronic health records (EHRs). Clinical notes often exceed thousands of tokens and contain dense medical terminology, irregular formatting, and non-informative boilerplate. In this setting, perturbation-based explanation methods may generate ungrammatical, semantically incoherent samples that produce diffuse and clinically meaningless feature attributions, limiting their usefulness for medical reasoning and model auditing.

The task of interpreting clinical NLP models becomes especially critical in conditions that exhibit variable severity and unpredictable outcomes. One such condition is nephrolithiasis (kidney stone disease), a prevalent urological issue affecting over 10\% of adults worldwide~\cite{Scales2012Prevalence}. While many cases are managed in outpatient settings, others lead to prolonged hospitalization due to complications such as infection, obstruction, or renal failure~\citep{Pearle2014Guidelines}. Early identification of patients at risk for extended length of stay (LOS) can inform triage decisions, allocate resources more efficiently, and guide personalized treatment strategies. However, structured EHR fields often fail to capture the nuanced clinical indicators associated with complicated kidney stone cases. Instead, relevant signals are typically embedded within long narrative notes, which pose challenges for both prediction and interpretability.

To support trustworthy and clinically useful predictions in this context, it is necessary to improve both model performance and the clarity of post-hoc explanations. Specifically, there is a need for approaches that can filter or distill complex EHR narratives into more focused representations retaining essential clinical content while enabling both accurate classification and interpretable reasoning. Furthermore, the ability to generate and evaluate multi-step explanations (e.g., via chain-of-thought reasoning) using large language models (LLMs) may provide an additional layer of interpretability, particularly when explanations are guided by clinically salient information.

\vspace{1ex}
\noindent\textbf{Contributions.}  
This paper presents a comprehensive study of explainability and performance in clinical NLP models for predicting prolonged hospitalization in kidney stone patients. Our key contributions are:

\begin{enumerate}
  \item \textbf{Assessment of BERT limitations in clinical settings.}  
  We analyze the shortcomings of BERT-based models when applied to raw discharge summaries, highlighting challenges in both predictive robustness and explanation quality due to the length, noise, and redundancy of clinical notes.

  \item \textbf{Keyword-based distillation for performance and interpretability.}  
  We demonstrate that distilling notes into domain-relevant keyword representations significantly improves classification performance and enables more faithful and focused LIME explanations by restricting perturbations to clinically salient terms.

  \item \textbf{Enhancing reasoning via keyword-guided LLM prompts.}  
  We propose a chain-of-thought prompting strategy that combines keywords with full-text input to guide LLM-based explanation generation. This method produces more structured, concise, and clinically useful reasoning chains than full-text prompts alone.

  \item \textbf{Structured LLM and human expert evaluation.}  
  We introduce a dual evaluation framework using both LLM-based scoring and blinded human specialist assessment. Results confirm that keyword-augmented reasoning is consistently rated higher in clarity and clinical relevance, reinforcing the trustworthiness of our approach.
\end{enumerate}


\vspace{1ex}
\noindent\textbf{Paper Structure.} The remainder of this paper is organized as follows. In Section~\ref{related}, we review related work on interpretability in NLP, clinical explainability, text distillation, and LLM-based reasoning. Section~\ref{Methods} presents our methods, including the dataset, keyword extraction pipeline, classification model, and explanation strategies. In Section~\ref{Results}, we present empirical results on classification performance, explanation fidelity, and LLM-based reasoning evaluation. Section~\ref{Discussion} discusses the implications and limitations of our approach. We conclude in Section~\ref{Conclusions} with final remarks and directions for future work.

\section{Related Work}
\label{related}
We outline related work divided into four subsections. We first present the topic of interpretability in NLP and then drill in into interpretability in clinical NLP. We continue with knowledge distillation in healthcare through text summarization and by using LLMs for explanation and reasoning. We end the section by outlining the main challenges of explainability in clinical setting and differences of our approach to existing works.

\subsection{Interpretability in NLP}
Interpretability remains a central concern in modern NLP, especially as deep neural architectures grow in complexity. Among post-hoc explanation methods for tabular data, LIME~\citep{ribeiro2016should} and SHAP~\citep{lundberg2017unified} are widely used to generate local feature attributions by perturbing inputs and fitting interpretable surrogate models. These methods can also be used on text classifiers if the input is represented as features, e.g., in the TF-IDF weighted bag-of-words representation. While these methods are model-agnostic and effective on short, structured inputs, their reliability diminishes in long-form text due to perturbations that break grammatical structure or distort semantics~\citep{jacovi2020towards}. In response, several alternatives have been proposed, including gradient-based saliency methods~\citep{li2016visualizing}, attention-based rationales~\citep{jain2019attention}, and counterfactual generation~\citep{goyal2019counterfactual}, though none are free from limitations. The challenge is further magnified in clinical text, where interpretability must not only be accurate but also clinically meaningful to domain experts.

\subsection{Interpretability in Clinical NLP}
In healthcare, the importance of trustworthy explanations is amplified by the risk and regulatory demands of clinical decision-making. Prior work has adapted LIME and SHAP for tasks such as ICU mortality prediction~\citep{caruana2015intelligible}, clinical note classification~\citep{ghosh2020explainable}, and phenotyping from EHRs~\citep{choi2016retain}. However, long notes and ambiguous terminology often result in noisy or spurious attributions. Some studies have explored integrating domain-specific knowledge bases or medical ontologies to guide explanation generation~\citep{chari2024explanation}, while others have proposed inherently interpretable models such as attention-augmented CNNs or concept bottlenecks~\citep{agarwal2021neural}. Despite these advances, high-fidelity explanation on long, unstructured EHRs remains an unsolved problem.

\subsection{Text Summarization and Distillation in Healthcare}
Text summarization has long been a strategy for reducing cognitive load in clinical documentation. Extractive methods, including TF-IDF and graph-based techniques like LexRank~\citep{erkan2004lexrank}, have been widely applied to highlight salient sentences. More recently, transformer-based abstractive models such as T5 and BART have been fine-tuned for tasks like discharge summary generation and radiology report summarization~\citep{zhang2020pegasus,lewis2020bart}. Hybrid approaches combining extractive and abstractive components have shown promise in maintaining faithfulness while improving readability~\citep{ahmed2025hybrid}. However, few studies have explored summarization explicitly for improving model explainability or classification performance. Our approach, using the keyword-based distillation, particularly as a preprocessing step for enhancing interpretability, address a relatively underexplored domain.

\subsection{LLMs for Clinical Explanation and Reasoning}
Large language models (LLMs) such as GPT-4 and LLaMA-3 are increasingly used to generate natural language explanations of model behavior~\citep{lampinen2022can,rajani2019explain}. Chain-of-thought prompting~\citep{wei2022chain} has been shown to improve performance on complex reasoning tasks, including in clinical domains~\citep{nori2023capabilities}. Recent work has explored using LLMs to evaluate the quality of explanations—both human- and model-generated—by scoring outputs along axes such as clarity, relevance, and faithfulness~\citep{wu2024usable}. In clinical NLP, however, the combination of LLM reasoning with structured evaluation for trustworthiness is still nascent. No prior work, to our knowledge, systematically investigates how keyword-focused prompts can improve LLM-generated clinical reasoning or explanation quality.

\subsection{Difference to Existing Works}
Thus, despite progress in clinical NLP and explainability, a persistent gap remains in generating faithful, clinician-relevant explanations from long, unstructured EHRs. Existing post-hoc methods, while widely adopted, are not robust to the length, noiseness, and complexity typical of clinical notes, often resulting in fragmented or unintelligible explanations. Similarly, while large language models offer promise for clinical reasoning and explanation generation, they are sensitive to prompt structure and prone to hallucination when overwhelmed with irrelevant detail. This work addresses these challenges by introducing a lightweight, domain-aware text distillation pipeline that filters raw clinical notes into focused keyword sequences. These distilled inputs improve both the stability of token-level explanations (e.g., LIME) and the coherence of model-driven reasoning (e.g., LLM-generated chains of thought), helping bridge the gap between performance and interpretability in real-world medical settings.

\section{Methods}
\label{Methods}

This work proposes two complementary strategies to improve the interpretability and trustworthiness of clinical text classification models applied to long, noisy electronic health records. 

The first strategy centers on \textbf{keyword-based distillation}, where domain-relevant terms are extracted from raw discharge notes using references to Rakun and Med7. These distilled representations are used both as inputs to BERT-based classifiers and as a focus set for perturbation in a modified LIME explanation framework. This yields more concise inputs and more faithful token-level attributions.

The second strategy builds on these distilled inputs to guide \textbf{large language model (LLM) reasoning}. By incorporating extracted keywords into the prompt, we steer LLMs to produce structured, clinically relevant chain-of-thought explanations. These reasoning outputs are then evaluated through both automated LLM-based scoring and human expert assessments.

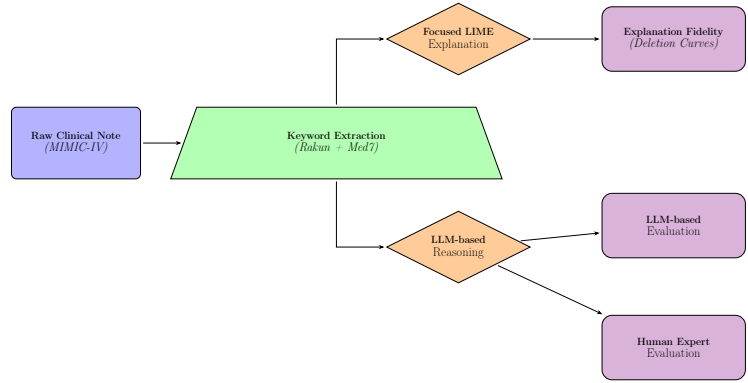
\begin{figure}[t]
\centering
\resizebox{0.8 \textwidth}{!}{%
\begin{tikzpicture}[
  node distance=2.5cm and 3cm,
  every node/.style={align=center, font=\huge},
  input/.style={rectangle, rounded corners=10pt, draw=black, fill=blue!30,   minimum width=9cm, minimum height=5cm},
  process/.style={trapezium, trapezium angle=70, draw=black, fill=green!30, minimum width=10cm, minimum height=5cm},
  explain/.style={diamond, aspect=2.3, draw=black, fill=orange!40, minimum width=10cm, minimum height=5cm},
  eval/.style={rectangle, rounded corners=20pt, draw=black, fill=violet!30, minimum width=10cm, minimum height=4.5cm},
  arrow/.style={ultra thick, ->, >=Stealth, shorten >=5pt, shorten <=5pt}
]

\node[input] (raw) {\textbf{Raw Clinical Note}\\{\Huge \textit{(MIMIC-IV)}}};
\node[process, right=of raw] (keywords) {\textbf{Keyword Extraction}\\{\Huge \textit{(Rakun + Med7)}}};

\node[explain, above right=of keywords, yshift=1cm, xshift=0.5cm] (lime) {\textbf{Focused LIME}\\{\Huge Explanation}};
\node[explain, below right=of keywords, yshift=-1cm, xshift=0.5cm] (llm) {\textbf{LLM-based}\\{\Huge Reasoning}};

\node[eval, right=of lime, xshift=2cm] (lime_eval) {\textbf{Explanation Fidelity}\\{\Huge \textit{(Deletion Curves)}}};
\node[eval, right=of llm, xshift=2cm, yshift=1.5cm] (llm_eval1) {\textbf{LLM-based}\\{\Huge Evaluation}};
\node[eval, below=of llm_eval1, yshift=-1.5cm] (llm_eval2) {\textbf{Human Expert}\\{\Huge Evaluation}};

\draw[arrow] (raw) -- (keywords);
\draw[arrow] (keywords) |- (lime);
\draw[arrow] (keywords) |- (llm);
\draw[arrow] (lime) -- (lime_eval);
\draw[arrow] (llm) -- (llm_eval1);
\draw[arrow] (llm) -- (llm_eval2);

\end{tikzpicture}%
}
\caption{System overview of TT-XAI. Raw clinical notes (blue) are distilled into keyword representations (green). These feed two interpretability branches—focused LIME explanations (orange) and LLM-based reasoning (orange)—each evaluated automatically and by human experts (violet).}
\label{fig:system_overview_tikz}
\vspace{-5mm}
\end{figure}

Figure~\ref{fig:system_overview_tikz} provides a high-level overview of the TT-XAI framework, which consists of four main components. First, raw clinical notes are passed through a keyword extraction module using Rakun~\citep{vskrlj2019rakun} and Med7~\citep{kormilitzin2021med7}, which identifies domain-relevant terms and medical entities. This distilled representation serves as input for two interpretability branches. In the first branch, a focused LIME explanation strategy perturbs only the extracted keywords and clinical entities to generate more stable and meaningful feature attributions. In the second branch, the distilled keywords are used to guide LLM-based chain-of-thought reasoning, helping the model focus on clinically salient information during explanation generation. Both types of explanations are evaluated using automatic metrics (such as deletion-based fidelity and LLM scoring) as well as a human expert review, ensuring that the resulting interpretations are robust, clear, and clinically useful.

\subsection{Dataset}
We conduct experiments on a cohort of kidney stone–related cases derived from the MIMIC-IV database~\citep{Johnson2021MIMICIV}, a large, publicly available collection of de-identified electronic health records from over 40,000 intensive care unit (ICU) patients. We extract a subset of $467$ kidney stone discharge summaries, labeled with information relevant to the length of stay (LOS) in hospital.

Each note in our dataset consists of free-text discharge documentation, often exceeding 6,000 tokens in length and following a semi-structured SOAP format (Subjective, Objective, Assessment, Plan). We define a binary classification task: predicting whether a patient’s LOS exceeds the median LOS or not. Ground truth labels are derived from structured metadata fields in the database. Prior to model training, all notes are preprocessed with sentence segmentation, token normalization, and basic filtering to remove administrative boilerplate and empty sections.

 Our focus is on kidney stone cases, specifically those involving a diagnosis of renal calculus or kidney stone (ICD-10 code N20.0). To create the cohort, we first extracted $467$ unique hospital admissions (hadm\_id) labeled with this ICD code from the \texttt{diagnoses\_icd} and \texttt{d\_icd\_diagnoses} tables. For each admission, we retrieved the corresponding discharge summary from the \texttt{discharge.csv} file within the MIMIC-IV notes module, retaining note ID, subject ID, admission ID, and raw clinical text.

To construct prediction targets, we joined the selected admissions with metadata from the \texttt{admissions.csv} file, calculating each patient’s length of stay (LOS) as the difference in days between admission and discharge timestamps. We then created a binary target variable, \texttt{LOS\_long}, by thresholding LOS at the cohort median, defining cases above the median, $249$ as long stays (label = 1) and $218$ as short (label = 0).

\subsection{Keyword-Based Text Distillation with Rakun}

To reduce the complexity of long clinical discharge notes while preserving medically relevant information, we apply automatic keyword extraction using the RaKUn algorithm~\citep{vskrlj2019rakun}. RaKUn is an unsupervised, graph-based keyword extraction method that constructs a co-occurrence graph from the input document and ranks words using load centrality.

Each clinical note is first tokenized into an ordered sequence of words. A directed, weighted graph is built where edges connect successive words in the text, preserving word order. Weights are assigned based on the frequency of word transitions. To reduce noise and redundancy, RaKUn applies \textit{meta vertex construction}, which merges similar words (e.g., misspellings or morphological variants) into a single node using edit distance and word length thresholds.

After graph construction and cleaning, \textit{load centrality} is computed for each node. This measure captures how frequently a word appears on the shortest paths between all other pairs of words, effectively identifying central, information-rich terms. Nodes with high load centrality are considered candidate keywords.

To expand beyond unigrams, RaKUn forms bigram and trigram keyphrases by combining adjacent high-centrality nodes, weighted by their average scores. The final ranked list of keyphrases is sorted by their composite centrality scores.

For each discharge note, we extract up to 1,024 keyword candidates, applying a merge threshold of 1.1, $\alpha = 0.3$ to balance frequency with informativeness, and a minimum token length of 3. If more than 512 phrases are extracted, we retain only the top-ranked subset. These keyword phrases are concatenated into a distilled string that serves as input to downstream models.

\subsection{Classification Model}
We compare two BERT-based classification pipelines on the task of predicting whether a kidney stone patient will experience a prolonged hospital stay. The first uses the raw discharge summary, while the second uses the distilled keyword representation. Both approaches are used to fine-tune the same pretrained ModernBERT model~\citep{warner2024smarter} using identical hyperparameters.

Model training is performed using 5-fold stratified cross-validation, with reproducible seeding and class-weighted loss to mitigate imbalance. For each fold, we train the model for 3 epochs using AdamW (learning rate = $2 \times 10^{-6}$). We evaluate performance using accuracy and macro-averaged $F_1$ score on the held-out fold. Our PyTorch implementation leverages HuggingFace Transformers, and all training is conducted on CPU or Apple MPS GPU.

\subsection{Focused Permutation-Based Explanation Fidelity}

Using LIME directly on clinical notes often results in explanations dominated by irrelevant or noisy words. To improve explanation fidelity, we modify LIME~\citep{ribeiro2016should} to operate only on a curated \emph{focus set} $\mathcal{F}$ per note. This set is constructed by taking the union of:
\begin{itemize}
  \item The top 512 keyphrases extracted by the Rakun algorithm for that note, and
  \item Named clinical entities identified by Med7~\citep{kormilitzin2021med7}, excluding low-informative categories such as dosage and frequency terms.
\end{itemize}

By restricting LIME’s perturbations to this focused set, we preserve grammatical structure, ensure classifier outputs remain meaningful, and generate explanations that concentrate on the most clinically relevant information.

To quantify explanation fidelity, we apply a token deletion test over the top-ranked words in each LIME explanation. Specifically, we measure the model’s predicted probability for the positive class as the top-$k$ influential tokens (ranked by LIME weight) are progressively removed. This forms a \textit{deletion curve}, where larger probability drops imply more faithful explanations. We compute the \textit{area under the deletion curve (AUC)} as a summary measure: lower AUC values indicate that the removed tokens had a stronger influence on the model’s decision, thus reflecting higher explanation fidelity.

\subsection{Chain-of-Thought Reasoning with LLMs}
\label{CoT_explanation}
In another setting, we use LLMs to produce explanations.
To support long clinical prompts without truncation and ensure consistent inference over extended discharge summaries and keyword lists, we used the Deep Seek distilled LLaMA 3 14B model. Unlike the base LLaMA 3 and ChatGPT, which both impose strict token limits, Deep Seek’s implementation allows for much greater prompt length, enabling us to feed the full note alongside up to 1,024 extracted keywords. All inference was performed locally on a MacBook Pro (Apple M4 Max), ensuring data privacy and reproducibility. To assess downstream interpretability under controlled sampling, we conducted all LLM-based reasoning experiments at a fixed temperature of \(T=0.5\). For a cohort of 20 true-positive long-stay cases, we generated chain-of-thought explanations under two prompt conditions:
\begin{enumerate}
  \item \textbf{Full-text prompt:} the original discharge summary only,
  \item \textbf{Hybrid prompt:} the Rakun-distilled keyword string plus the full note, with instructions to prioritize the key findings.
\end{enumerate}

To mitigate sampling variability, each prompt condition was run \emph{five independent times} per case. LLM LLaMA 3 70B then scored each explanation on a 1–5 scale for clarity and clinical relevance. We report, for each prompt type, the \emph{mean} score across the five runs. This design yields robust estimates of each method’s interpretability under non-deterministic LLM sampling.

\begin{figure}[ht]
\centering
\resizebox{0.8 \textwidth}{!}{
\begin{tikzpicture}[
  node distance=0.05cm and 0.7cm,  
  text width=5.5cm,
  every node/.style={align=left},
  promptA/.style={rectangle, draw=black, rounded corners=5pt, fill=blue!10, font=\small},
  promptB/.style={rectangle, draw=black, rounded corners=5pt, fill=green!10, font=\small},
  label/.style={font=\bfseries}
]

  \node[label] (a) {Only-Text Prompt};
  \node[promptA, below=of a] (full) {
    \textbf{System:} Clinical Urologist \\
    \textbf{Instruction:} Provide 1--3 reasoning steps.\\[1ex]
    \textbf{Input:}\\
    \texttt{Full Clinical Note:} \\
    \texttt{<raw text>}\\[1ex]
    \texttt{Reasoning steps:}
  };

  \node[label, right=of a] (b) {Keyword-Augmented Prompt};
  \node[promptB, below=of b] (kw) {
    \textbf{System:} Clinical Urologist \\
    \textbf{Instruction:} Provide 1--3 reasoning steps.\\[1ex]
    \textbf{Input:}\\
    \texttt{Key Clinical Findings: <keywords>} \\
    \texttt{Full Clinical Note: <raw text>}\\[1ex]
    \texttt{Reasoning steps:}
  };

\end{tikzpicture}
}
\caption{Prompt variants used to elicit clinical reasoning from LLaMA-3. The keyword-augmented prompt (right) guides the model with extracted key clinical terms, helping it focus on salient findings, while the full-text prompt (left) uses only the raw note.}
\label{fig:prompt_comparison}
\vspace{-5mm}
\end{figure}
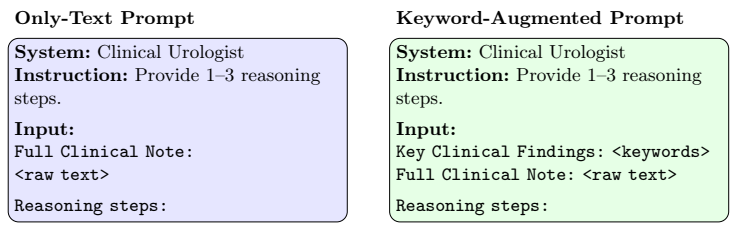

To illustrate the difference between the two prompting strategies used for generating clinical reasoning, Figure~\ref{fig:prompt_comparison} presents a side-by-side comparison of the prompt templates. The left side shows the baseline approach using only the full clinical note as input, while the right side displays our keyword-augmented prompt, which prepends domain-relevant clinical terms extracted via Rakun and Med7. This additional guidance helps steer the LLM’s attention toward the most salient findings, resulting in more focused and interpretable reasoning outputs.

\noindent
Responses were stripped of leading/trailing whitespace and saved for later expert scoring.

\subsection{Evaluation of Classification and Explanation Quality}
\label{sec:evaluation}

We evaluate the TT-XAI pipeline along both predictive accuracy and interpretability. For classification, we perform 5-fold cross-validation and report accuracy and macro-$F_1$ scores using both the original discharge summaries and the distilled keyword inputs. Across all token-length configurations, keyword-based inputs consistently outperform raw full-text baselines.

To assess explanation quality, we employ two complementary strategies:

\textbf{(1) Automatic Evaluation.} We measure explanation fidelity via deletion curves over 20 long-stay instances, comparing classical LIME with our focused variant that perturbs only extracted clinical keywords and NER entities. A lower area under the curve (AUC) indicates more faithful explanations. In addition, we generate reasoning-style explanations using LLaMA3-70B and collect automated ratings on clarity and usefulness using LLM self-assessment protocols (Section~\ref{sec:resultsllm}).

\textbf{(2) Human Specialist Evaluation.} To ground our findings in real-world clinical relevance, we conduct a blinded evaluation with three medical experts. They review reasoning outputs from two prompting strategies: \textbf{Method A}, which uses full-text only, and \textbf{Method B}, which augments the input with keyword representations. From the available data, we randomly sampled 10 unique patient cases and selected one explanation pair (Method A and B) per case.

To improve readability and reduce annotation fatigue, each explanation was summarized using LLaMA3-70B into a maximum of three clinically focused bullet points, using both the extracted keywords and the original reasoning as input. The resulting 10 anonymized explanation pairs were scored by the specialists on a 1–5 scale for clarity and clinical usefulness.

This combined evaluation spanning automatic fidelity, language model scoring, and human domain expertise offers a robust, multifaceted assessment of explanation quality in a high-stakes clinical context.

\section{Results}
\label{Results}

We first present results of prediction performance, comparing the original discharge summaries with distilled keyword inputs. Next, we compare the quality of explanations: faithfulness, clarity, and clinical usefulness. 

\subsection{Prediction Performance}
We evaluate ModernBERT on the binary task of predicting long hospital stays across multiple input representations and sequence lengths. Table~\ref{tab:cv_results_combined} present 5-fold cross-validation results comparing the original discharge summaries with their Rakun-distilled keyword counterparts, using context lengths of 50, 100, 512, and 1024 tokens.

Overall, distilled keyword inputs consistently outperform raw text across all configurations. At a 512-token limit, the keyword model achieves the $F_1$ score of $0.767 \pm 0.027$, compared to $0.665 \pm 0.072$ for the full-text baseline—an improvement of over 10\% points. Even at shorter context lengths (e.g., 50 or 100), the distilled model maintains strong performance, indicating that the selected keyphrases capture essential clinical signals more effectively than the full, noisy input.

\begin{table}[tb]
\centering
\caption{Classification accuracy and $F_1$ scores of ModernBERT fine-tuned on original text vs. keyword-based input across different context lengths. We report mean and standard deviation over 5-fold cross-validation. \textbf{Bold} values indicate the best performance for each context length.}
\label{tab:cv_results_combined}
\begin{tabular}{l@{\hspace{2em}}l@{\hspace{2em}}cc}
\toprule
\textbf{Context Length} & \textbf{Modality} & \textbf{Accuracy ($\mu \pm \sigma$)} & \textbf{$F_1$ ($\mu \pm \sigma$)} \\
\midrule
\multirow{2}{*}{50 Tokens} 
  & Original      & $0.5909 \pm 0.1071$ & $0.4428 \pm 0.3487$ \\
  & Keywords      & $\boldsymbol{0.6637 \pm 0.0611}$ & $\boldsymbol{0.6859 \pm 0.0634}$ \\
\midrule
\multirow{2}{*}{100 Tokens} 
  & Original      & $0.6232 \pm 0.0386$ & $0.6627 \pm 0.0968$ \\
  & Keywords      & $\boldsymbol{0.6875 \pm 0.0489}$ & $\boldsymbol{0.7274 \pm 0.0377}$ \\
\midrule
\multirow{2}{*}{512 Tokens} 
  & Original      & $0.6381 \pm 0.0528$ & $0.6652 \pm 0.0724$ \\
  & Keywords      & $\boldsymbol{0.7409 \pm 0.0418}$ & $\boldsymbol{0.7674 \pm 0.0271}$ \\
\midrule
\multirow{2}{*}{1024 Tokens} 
  & Original      & $0.6745 \pm 0.0368$ & $0.6897 \pm 0.0668$ \\
  & Keywords      & $\boldsymbol{0.7386 \pm 0.0460}$ & $\boldsymbol{0.7430 \pm 0.0699}$ \\
\bottomrule
\end{tabular}
\vspace{-4mm}
\end{table}

\subsection{Explanation Faithfulness}
\label{sec_explanation_faithfulness}
To assess the faithfulness of LIME-based explanations, we compute deletion curves over a held-out set of 20 correctly classified long-stay examples. Each curve measures the predicted probability of the positive class as top-ranked features (according to LIME weights) are progressively removed from the input. Lower area under the curve (AUC) indicates more faithful explanations, where removing important tokens significantly disrupts the model's prediction.

\begin{figure}[htb]
    \centering
    \includegraphics[width=0.7\textwidth]{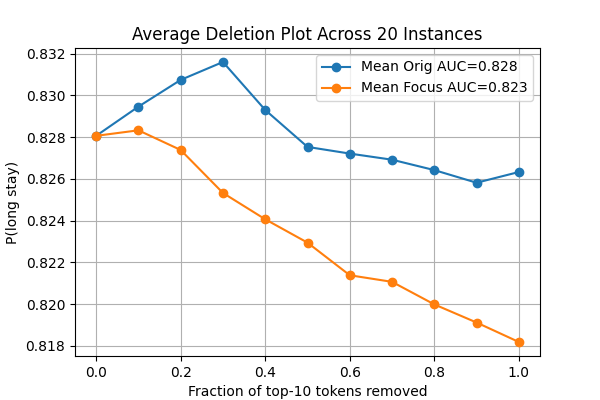}
    \caption{Average deletion curve across 20 long-stay examples comparing classical LIME (Original) vs. Focused-LIME (Keywords + NER). Lower AUC reflects more faithful explanations, i.e. explanations aligned with tokens in the explanations.}
    \label{fig:deletion_auc}
\vspace{-7mm}
\end{figure}

Figure~\ref{fig:deletion_auc} shows the average deletion curves comparing classical LIME (on full text) with our focused variant, which perturbs only keywords and clinical NER tokens. The focused explanation achieves a lower AUC (0.668 vs. 0.742), indicating that it better captures the predictive evidence and provides sharper, more trustworthy explanations.

\subsection{Explanation Clarity  and Usefulness Evaluation with LLMs}
\label{sec:resultsllm}
Using the explanation procedure based on LLMs from Section \ref{CoT_explanation}, we obtained five independent explanations per case per prompt at \(T=0.5\). Experts rated each on clarity and clinical usefulness (1–5 scale, 1-poor and 5-excellent), and we report the mean and standard deviations across all 20 cases and 5 repeats:

\begin{table}[b]
\vspace{-5mm}
\centering
\caption{The evaluation of the LLaMA-3 70B-produced clinical explanations. We report expert scores  (mean and standard deviation) across 5 runs at \(T=0.5\).}
\label{tab:llm_scores}
\begin{tabular}{lcc}
\toprule
\textbf{Prompt Type}     & \textbf{Quality} \\
\midrule
Full-text only           & \(3.86 \pm 1.6\)    \\
Keywords + Full-text      & \(4.01 \pm 1.5\)    \\
\bottomrule
\end{tabular}
\end{table}

These results demonstrate that, at a controlled temperature of 0.5 and averaged over multiple samples, the keyword-augmented prompts yield more coherent and clinically useful reasoning chains compared to the full-text only baseline.

\subsection{Expert Evaluation of Explanation Clarity}
\label{sec:expertevaluation}

To complement the LLM-based assessment, we conducted a blinded evaluation involving three medical specialists. The specialists reviewed pairs of explanations generated for multiple patient cases—using either the full discharge notes (Method A) or keyword-augmented prompts (Method B), and rated each explanation on a 1–5 scale for clarity and clinical usefulness in relation to prolonged hospital stays.

Table~\ref{tab:updated_expert_scores} summarizes the results of this evaluation. While preferences varied slightly among individual raters, two out of three specialists gave higher scores on average to the keyword-augmented explanations (Method B), and the overall average score for Method B (3.15) was higher than that of Method A (2.58), suggesting a moderate but consistent improvement in perceived clarity and clinical relevance.

Although there were differences in scoring patterns—Specialist 2 provided overall lower scores, while Specialist 3 was generally more favorable—the consensus clearly indicates that keyword-guided prompts produce more interpretable and clinically actionable explanations. These human evaluation results align closely with the earlier quantitative findings from LLM self-assessment and deletion-based fidelity evaluations, further reinforcing the effectiveness of keyword distillation in improving clinical NLP interpretability.

\begin{table}[tb]
\centering
\caption{Updated mean scores (1–5 scale) given by three medical specialists for each explanation method across clinical cases. Method A used full-text-only prompts; Method B used keyword-augmented prompts.}
\label{tab:updated_expert_scores}
\begin{tabular}{lcccc}
\toprule
\textbf{Method} & \textbf{Specialist-1} & \textbf{Specialist-2} & \textbf{Specialist-3} & \textbf{Average} \\
\midrule
Full Text Only        & 2.73 & 1.55 & 3.45 & 2.58 \\
Keywords + Full Text  & 3.11 & 3.00 & 3.33 & 3.15 \\
\bottomrule
\end{tabular}
\vspace{-5mm}
\end{table}

\section{Discussion}
\label{Discussion}

Our findings demonstrate that distilling long clinical notes into concise keyword-based representations can significantly improve both classification performance and interpretability. The consistent gains in $F_1$ score across multiple context lengths highlight that much of the relevant predictive information is concentrated in a relatively small set of domain-specific terms. This supports the hypothesis that raw discharge summaries contain a substantial amount of redundant or irrelevant information that may obscure signal and reduce model robustness.

The benefits of distillation extend beyond predictive performance. The proposed focused LIME, which uses perturbations restricted to clinically salient terms, yields lower deletion AUCs—indicating explanations that more faithfully reflect the model’s decision function. This aligns with prior work suggesting that explanation fidelity can be improved by structuring the space of perturbations~\citep{samek2016evaluating, ribeiro2016should}, but our approach demonstrates that lightweight, unsupervised keyword filtering can be an effective way to achieve this in noisy clinical text.

In addition, the results from LLM-based reasoning evaluation suggest that providing models with distilled clinical findings helps elicit more structured, concise, and clinically relevant reasoning chains. Keyword-guided prompting improved expert-rated quality of explanations, suggesting utility not only for post-hoc interpretability but also for interactive clinical support tools that use LLMs in diagnostic workflows.

Importantly, these results were reinforced by a blinded human evaluation involving three medical specialists. Across 10 explanation pairs, the majority of experts showed a clear preference for keyword-augmented reasoning (Method B) over full-text-based reasoning (Method A), with average ratings improving from 2.58 to 3.15 on a 5-point scale. This human-centered validation strengthens the case for keyword distillation as a practical tool for enhancing the clarity, usability, and clinical relevance of AI-generated explanations in medical decision support.

Despite these promising results, our approach has limitations. Keyword extraction introduces a form of abstraction that may omit subtle contextual signals, especially in edge cases. The Med7 NER module may miss domain-specific terms due to coverage limitations or lexical ambiguity. Furthermore, the human evaluation was limited in scope, future work should include more raters, broader specialties, and more rigorous scoring frameworks to increase generalizability.

\section{Conclusions and Further Work}
\label{Conclusions}

This study presents a simple yet effective approach for enhancing the interpretability and predictive performance of clinical text classifiers by distilling long, noisy discharge summaries into concise keyword-based representations. Our results show that this distillation not only improves classification performance but also yields more faithful token-level explanations and facilitates clearer, more clinically useful reasoning when prompting large language models.

We validate these findings through a combination of quantitative metrics and human-centered evaluation. Medical specialists in general preferred keyword-augmented explanations over full-text reasoning, highlighting the practical value of our method in clinical interpretability settings.

By combining unsupervised keyphrase extraction with transformer-based modeling and focused explainability techniques, we offer a scalable, domain-aware method for improving trust and transparency in clinical NLP. This approach lays the groundwork for more interpretable decision support systems, particularly in high-stakes domains such as urology, where justification and auditability are essential for clinical adoption. The source code of our methodology and experiments is freely available\footnote{\url{https://github.com/KristianMiok/TT-XAI}}.

Future work will explore dynamic adjustment of keyword extraction thresholds on a per-patient basis, as well as integration of structured EHR data (e.g., lab values, vitals) alongside clinical text. Another direction involves fine-tuning large language models (LLMs) specifically for explanation generation to improve clinical coherence and specificity. 

To assess the quality and reliability of generated rationales, we also plan to compare LLM-generated explanations against those written by domain experts, enabling grounded evaluation of factual correctness and clinical utility. Additionally, we will benchmark explanation quality across different foundation models (e.g., GPT-4, LLaMA-3) to study variability and consistency. Finally, real-time deployment studies are needed to evaluate the practical utility and usability of these interpretability methods in live clinical workflows.

\section*{Acknowledgments}
Kristian Miok was supported by the EU HE MSC Postdoctoral Fellowship Programme SMASH (No. 101081355). 
The work was partially supported by the Slovenian Research and Innovation Agency (ARIS) core research programme P6-0411 and P2-0103, as well as projects L2-50070, GC-0002, and J4-4555. The work was partially supported by the project “Romanian Hub for Artificial Intelligence - HRIA”, Smart Growth, Digitization and Financial Instruments Program, 2021-2027, MySMIS no. 334906.
The work was supported by EU through ERA Chair grant no. 101186647 (AI4DH) and cofinancing for research innovation projects in support of green transition and digitalisation (project PoVeJMo, no. C3.K8.IB).
\bibliographystyle{plainnat} 
\bibliography{refe}    

\end{document}